# PTRL: Prior Transfer Deep Reinforcement Learning for Legged Robots Locomotion

Haodong Huang, Shilong Sun, Zida Zhao, Hailin Huang,
Changqing Shen, Wenfu Xu

*Abstract*—In the field of legged robot motion control, reinforcement learning (RL) holds great promise but faces two major challenges: high computational cost for training individual robots and poor generalization of trained models. To address these problems, this paper proposes a novel framework called Prior Transfer Reinforcement Learning (PTRL), which improves both training efficiency and model transferability across different robots. Drawing inspiration from model transfer techniques in deep learning, PTRL introduces a fine-tuning mechanism that selectively freezes layers of the policy network during transfer, making it the first to apply such a method in RL. The framework consists of three stages: pre-training on a source robot using the Proximal Policy Optimization (PPO) algorithm, transferring the learned policy to a target robot, and fine-tuning with partial network freezing. Extensive experiments on various robot platforms confirm that this approach significantly reduces training time while maintaining or even improving performance. Moreover, the study quantitatively analyzes how the ratio of frozen layers affects transfer results, providing valuable insights into optimizing the process. The experimental outcomes show that PTRL achieves better walking control performance and demonstrates strong generalization and adaptability, offering a promising solution for efficient and scalable RL-based control of legged robots.

*Index Terms*—Legged Robots, Transfer Learning, Deep Reinforcement Learning

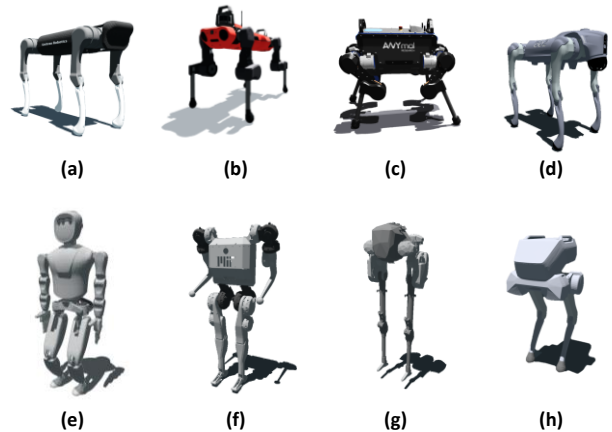

Fig. 1. Demonstrates the application of our PTRL-based method in full-body control across various types of robots. (a) A1, (b) AnymalC, (c) AnymalB, (d) Go2, (e) XBot-L humanoid, (f) MIT humanoid, (g) Cassie, (h) LimX Dynamics TRON1.

## I. INTRODUCTION

DRIVEN by rapid advancements in motor technology, legged robots have achieved significant progress in various aspects, particularly in locomotion capabilities. Legged robots possess load-bearing capabilities and can traverse various complex terrains, offering crucial support in manufacturing, construction, rescue operations, and other fields. The core competency of legged robots lies in the quality of their locomotion performance [1], [2], [3], [4], [5]. Consequently, substantial research has focused on developing motion-generation methods for legged robots to handle diverse and complex tasks. Traditional model-based approaches [6], such as Model Predictive Control (MPC) [7], Whole-Body Control (WBC) [8], and hybrid strategies combining the two [9], have achieved notable success in improving motion flexibility and stability. These achievements can be attributed to their rigorous theoretical foundations and precise modeling of robot-environment interactions [10], though they typically require extensive expertise and meticulous engineering adjustments.

Currently, deep reinforcement learning (DRL) methods are rapidly gaining traction, distinguished by their independence from precise physical models. By leveraging deep neural networks (DNNs) as powerful function approximators, DRL approaches can directly map raw state spaces to action spaces, enabling autonomous learning of optimal control strategies for complex tasks [11]. The introduction of domain randomization during training, which generates diverse data samples across varied environments, effectively bridges the gap between simulation and the real world, significantly enhancing the generalization capabilities of DRL algorithms. Furthermore, the sophisticated design of complex and flexible reward functions has driven breakthroughs in legged robot locomotion, achieving substantial progress in stability, efficiency, and adaptability. In addition, DRL harnesses parallel training techniques, markedly

This work was supported in part by State Key Laboratory of Robotics and System Under Grant KY511223001, in part by the National Natural Science Foundation of China under Grant 52475112, in part by Guangdong Basic and Applied Basic Research Foundation Under Grant 2024A1515012041. (*Corresponding author: Shilong Sun*.)

Haodong Huang, Zida Zhao, Shilong Sun, Chiyao Li, Hailin Huang and Wenfu Xu are with the School of Robotics and Advanced Manufacturing, Harbin Institute of Technology, Shenzhen , 518055, China (e-mails: hhd1340201839@163.com; sunshilong@hit.edu.cn a1320943099@163.com; huanghailin@hit.edu.cn; wfxu@hit.edu.cn ).

Changqing Shen is with the School of Rail Transportation, Soochow University, Suzhou 215006, China, and also with Suzhou Boyata Industrial Internet Company Ltd., Suzhou 215100, China (e-mail: cqshen@suda.edu.cn ).

Wenfu Xu is with the Key University Laboratory of Mechanism and Machine Theory and Intelligent Unmanned Systems of Guangdong Shenzhen, Guangdong 518055, China.

reducing training times and dramatically improving research efficiency. As a result, DRL has emerged as a dominant direction in the fields of intelligent control and robotics research [12], [13], [14].

Despite the ability of DRL methods to significantly enhance robot learning efficiency through parallel training, the training process for multiple heterogeneous robots remains computationally expensive and time-intensive. The introduction of complex terrains further escalates computational demands, posing greater challenges to the training process. In the field of deep learning, transfer learning has emerged as a pivotal technique aimed at transferring knowledge acquired in one domain (the source domain) to another related but distinct domain (the target domain). Its primary objective is to leverage information from the source domain to reduce the training complexity and improve the performance of models in the target domain [15]. Transfer learning has been widely applied in areas such as natural language processing, computer vision, and robotic control, demonstrating exceptional performance in cross-domain tasks and few-shot learning scenarios. Notably, it offers robust solutions to challenges posed by limited data and computational resources, making it an indispensable tool in addressing these constraints.

In practical applications of humanoid robot motion control, reinforcement learning is typically employed for training in simulation environments until the training converges, after which the policy is deployed to real robots via sim-to-real transfer. However, training a policy that can be directly deployed to an actual robot often requires a significant amount of time, particularly when multiple different robots need to be deployed, resulting in substantial computational resources and time costs. To address this issue, inspired by model transfer techniques in rotating machinery fault diagnosis to improve task efficiency [16] and fine-tuning methods from natural language processing (NLP) [17], this paper proposes a novel training paradigm for legged robots. This paradigm leverages prior knowledge by training an effective policy on a single-legged robot and subsequently transferring it to other robots, followed by parameter fine-tuning, thereby significantly reducing computational resource consumption and training time costs. Specifically, this paper combines DRL with actor model freezing techniques to propose a Prior Transfer Reinforcement Learning (PTRL) solution to tackle the high computational and time costs associated with retraining for different-legged robots. Initially, the Proximal Policy Optimization (PPO) algorithm is used to train a single-legged robot (such as a quadruped or humanoid robot) until a converged policy network is obtained. The network parameters of the actor part of this policy network are then transferred to another legged robot, with specific network layers frozen (i.e., set to be non-trainable), and only the unfrozen layers undergo parameter fine-tuning. Finally, by exploring different layer freezing configurations and combinations, the transfer performance is optimized, further validating the method's generalization capability and efficiency.

The main contributions of this study can be summarized as follows:

1) A novel-legged robot motion control method that combines transfer learning and reinforcement learning, leveraging prior knowledge to accelerate the robot training process.
2) The method's transferability and model generalizability were validated through cross-validation experiments conducted on quadruped robots and multiple humanoid robot platforms.
3) Compared to traditional reinforcement learning approaches, this study achieved reductions in computational resources and training time while achieving comparable results in walking tasks, thereby demonstrating the effectiveness of the proposed method.

The structure of this study is organized as follows: Section II provides a brief overview of the research progress in related fields, including DRL and transfer learning. Section III presents a detailed description of the proposed control framework, including the design of network hyperparameters and the training process. Section IV validates the proposed control model through experiments conducted on legged robots. Finally, Section V summarizes the main contributions of this study and discusses potential future research directions.

## II. RELATED WORKS

### A. RL-Based Methods

In recent years, RL has been widely applied to the motion control of legged robots. Yan et al. [18] designed a series of reward functions by referencing motion capture data, enabling the robot's motion to approximate the target trajectory gradually. Jeon et al. [19] benchmarked humanoid robots using both standard reward shaping methods and potential-based reward shaping (PBRS), finding that PBRS performed excellently in terms of stability and validated this on the MIT humanoid robot platform. Researchers have extensively explored reward function design, significantly improving the naturalness and smoothness of robot motion. To further optimize the stability and naturalness of robot motion, the Adversarial Motion Priors (AMP) imitation learning method [20] emerged. This approach can enable robot motion to resemble human walking more closely and has been successfully applied to real robots [21], [22]. Chen et al. [23] proposed a hybrid method that trains the upper body of humanoid robots using motion capture data while employing DRL to train the lower body. SERIFI et al. [24] introduced a Robot Motion Diffusion Model, which not only mimics human motion but also adheres to the robot's morphological and physical constraints. Moreover, to address the issue of the discrepancy between the simulation's extensive observation variables and the limited real-world observations, the Teacher-Student Strategy [25] offers an effective solution. This method first trains a teacher policy with privileged information, and then transfers the knowledge to a student policy via knowledge distillation, allowing the student policy to generate the same actions as the teacher policy using only intrinsic inputs. Additionally, the Concurrent Teacher-Student (CTS) method [26] further enhances policy learning efficiency and performance by simultaneously training both teacher and student policies, leading to excellent performance of legged robots in various tasks.

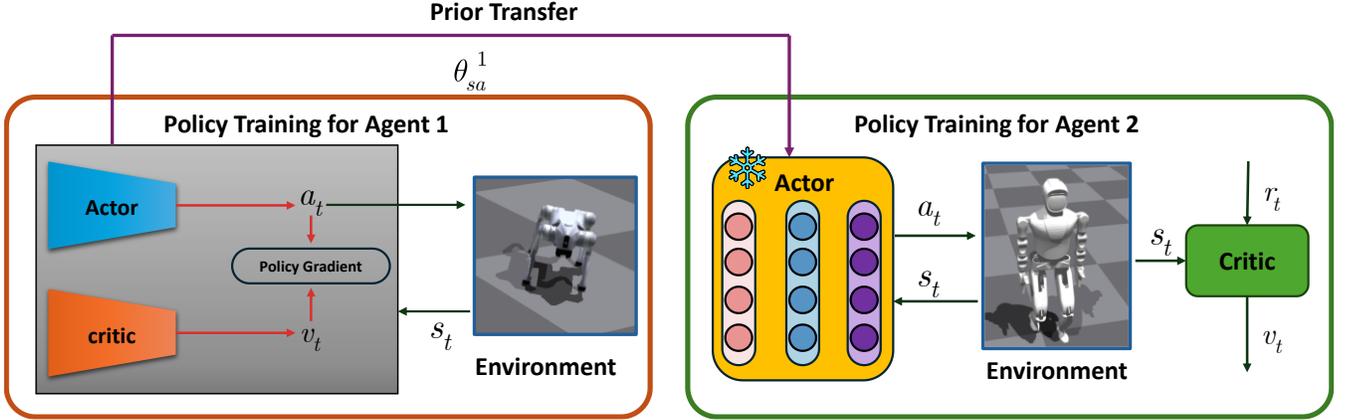

Fig. 2. An overview of PTRL includes two key modules: SLRT first trains the locomotion policy of a quadruped robot, then transfers the actor from the trained policy to a humanoid robot, freezing part of the layers to accelerate the humanoid's training.

*B. Prior Knowledge-Based Methods*

Transfer learning has been extensively applied across various domains of deep neural networks. In computer vision, it is widely used for tasks such as object detection [27] and medical image classification [28], significantly improving task accuracy and efficiency. In natural language processing, transfer learning finds applications in machine translation [29] and sentiment classification [30], where pre-trained models are leveraged for rapid adaptation to target tasks. In the context of legged robot control, Berseth et al. [31] employed knowledge distillation to incrementally enhance policies incrementally, enabling the handling of tasks with increased complexity. Qi et al. [32] addressed the issue of retraining caused by changes in models and reward functions during sim-to-real transfer by proposing a continuation-based optimization approach, which greatly reduces training time. Similarly, Smith et al. [33] introduced a framework that transfers the experience of existing controllers to new tasks, facilitating the rapid acquisition of complex skills.

## III. METHOD

This study aims to develop a training paradigm applicable to various legged robots, enabling models trained on one robot to serve as priors for training on another, thereby accelerating subsequent training processes. The proposed Prior Transfer Reinforcement Learning (PTRL) method combines model freezing with reinforcement learning to inherit and utilize prior knowledge of motion effectively. Fig. 2 illustrates the overall framework of PTRL, and the subsequent sections will provide a detailed explanation of its implementation and methodology.

*A. Preliminaries*

*RL for Legged Robots*: In the application of RL to legged robots, the problem is typically formulated as a partially observable Markov decision process (POMDP), represented by a six-tuple $\langle \mathcal{S}, \mathcal{O}, \mathcal{A}, \mathcal{R}, p, \gamma \rangle$. Here: $\mathcal{S}$ denotes the set of all possible states in the environment, $\mathcal{O} \subseteq \mathcal{S}$ represents the subset of observable states containing information accessible to the agent, $\mathcal{A}$ defines the set of all actions the robot can execute, $\mathcal{R}: \mathcal{S} \times \mathcal{A} \to \mathbb{R}$ assigns a scalar reward to each state-action pair, $p: \mathcal{S} \times \mathcal{A} \to \mathbb{P}(\mathcal{S})$ describes the probability distribution for transitioning from the current state to a new state given an action, $\gamma \in [0,1]$ is a discount factor balancing the influence of future rewards on current decisions. During the process, at each timestep $t$, the agent receives an observation $o_t \in \mathcal{O}$ from the environment and selects an action $a_t \in \mathcal{A}$ based on the policy $\pi(a_t | o_t)$. The environment then updates its state $s_{t+1}$ according to the transition probability $s_{t+1} \sim p(s_{t+1} | s_t, a_t)$, and the robot receives a reward $r_t = R(s_t, a_t)$. The objective of RL is to optimize the policy by maximizing the expected cumulative reward:

$$J(\theta) = \mathbb{E}_{\pi_\theta} \left[ \sum_{t=0}^{T-1} \gamma^t r_t \right] \quad (1)$$

In this study, we utilize the PPO algorithm to train the robot. PPO is an efficient policy optimization method that ensures stability and convergence by limiting the step size of policy updates. During each update, PPO introduces a clipping function to constrain policy changes, striking a balance between exploration and convergence. The objective of PPO can be formulated as follows:

$$\mathcal{L}_\pi = \min \left[ \frac{\pi(a_t | o_t)}{\pi_b(a_t | o_t)} A^{\pi_b}(o_t, a_t), \right.$$
$$\left. \text{clip}\left( \frac{\pi(a_t | o_t)}{\pi_b(a_t | o_t)}, 1-\varepsilon, 1+\varepsilon \right) A^{\pi_b}(o_t, a_t) \right] \quad (2)$$

Here, $\pi$ represents the target policy to be optimized, while $\pi_b$ denoting the behavior policy used for data sampling. The parameter $\varepsilon$ specifies the clipping range in PPO, which constrains the extent of policy updates to ensure stability during training.

*Transfer learning*: Transfer learning involves integrating and optimizing knowledge from a source domain and a target domain. A common approach is to minimize the distributional discrepancy between the source and target domains, often measured using Maximum Mean Discrepancy (MMD). This technique enhances the model's learning efficiency in the target domain. Suppose the dataset of the source domain is $D_s = \{(x_i, y_i)\}_{i=1}^{N_s}$, and that of the target domain is $D_t = \{(x_i, y_i)\}_{j=1}^{N_t}$, where $x_i, x_j$ denote the inputs of the source and target domains, respectively, and $y_i, y_j$ represent their corresponding labels. The objective of transfer learning is to minimize this discrepancy and integrate it into the training process via a loss function. The calculation is given by the following formula:

$$\mathcal{L}_{\text{total}} = \mathcal{L}_t + \lambda \cdot \text{MMD}(D_s, D_t) \tag{3}$$

In this study, we first train a policy $\pi(\theta_s)$ using a source robot, which $s$ denotes the source domain robot. The actor network in this policy is defined as $\pi(\theta_{sa})$, where $\theta_{sa} = \{\theta_{sa}^{(1)}, \theta_{sa}^{(2)}\}$ representing the fixed-layer parameters and trainable-layer parameters, respectively. For the target domain robot, the policy $\pi(\theta_t)$ adopts an actor model defined as $\pi(\theta_{ta})$, During model initialization, the network parameters of the source and target domains are transferred using the following approach:

$$\theta_{ta}^{(1)} = \theta_{sa}^{(1)}, \theta_{ta}^{(2)} = \theta_{sa}^{(2)} \tag{4}$$

Here, $\theta_{ta}^{(1)}$ is kept fixed, while $\theta_{ta}^{(2)}$ remaining trainable. This approach enables the actor network of the target domain robot to inherit knowledge from the source domain robot during initialization while retaining the ability to adapt to the specific requirements of the target domain robot.

The optimization objective for the target domain robot's policy is defined as:

$$J(\theta) = \mathbb{E}_{\pi_{(\theta_{ta}^{(2)}, \theta_{tc})}} \left[ \sum_{t=0}^{T-1} \gamma^t r_t \right] \tag{5}$$

where $\theta_{tc}$ represents the model parameters of the critic.

*B. Hardware Platform*

In this study, we validate the proposed method using several legged robots, including the Unitree Go2, MIT Humanoid, XBot-L, Unitree A1, Anymal_B, Anymal_C, LimX Dynamics TRON1, and Cassie. The Go2, A1, AnymalB, and AnymalC robots each have 12 degrees of freedom (DOF), with 3 DOF per leg: 2 DOF in the hip joint and 1 DOF in the knee joint. The MIT humanoid robot has 18 DOF in total. However, since the focus of this research is on leg motion, the shoulder movements are not explored in detail, and only 10 DOF corresponding to the legs are considered. For the XBot-L humanoid robot and Cassie, the leg DOF is used, totaling 12 DOF, while the LimX Dynamics TRON1 robot has 6 DOF in total. Detailed illustrations of the robot models can be found in Fig. 1.

*C. Overview of PTRL*

The overview of PTRL is shown in Fig. 2. Specifically, we use PTRL to sequentially train two different legged robots: the source domain robot and the target domain robot. This section will introduce the two modules of PTRL.

**SLRT** (Source Legged Robot Training): As shown in Fig. 2, training begins with the Go2 robot. The corresponding optimization process for SLRT is described in Eq. (1) and Eq. (2).

**TLRT** (Target Legged Robot Training): As shown in Fig. 2, the main objective of TLRT is to fine-tune the target robot's model by introducing the actor network parameters from SLRT, thus leveraging prior knowledge. The optimization process for this stage is described in Eq.(2) and Eq.(5).

*D. Training Procedure of PTRL*

The training process of PTRL is shown in Fig. 2. In Phase 1, the Go2 robot from Unitree is trained using reinforcement learning to obtain a walking policy, from which the actor network is extracted. In the second phase, a portion of the actor network is frozen, transferred directly to the humanoid robot, and fixed, while the remaining unfixed network parameters are trained. The pseudocode of the algorithm is shown in Algorithm 1.

---

**Algorithm 1** PTRL

**Stage 1**(SLRT): Initialize $\pi(\theta_{sa})$ and critic $\pi(\theta_{sc})$

    **for** $i \leftarrow 1$ to $L$ **do**

        $a_t = \pi(s_t; \theta_{sa})$

        Execute $a_t$, observe the next state $s_{t+1}$, reward $r_t$

        Update $\pi(\theta_{sc})$ using observed $(s_t, a_t, r_t, s_{t+1})$

        Update $\pi(\theta_{sa})$

    **end for**

**Stage 2** (TLRT): Copy $\pi(\theta_{sa})$ to initialize $\pi(\theta_{ta})$, initialize $\pi(\theta_{tc})$, Freeze $\theta_{ta}^{(1)}$

    **for** $i \leftarrow 1$ to $L$ **do**

        $a_t = \pi(s_t; \theta_{ta})$

        Execute $a_t$, observe the next state $s_{t+1}$, reward $r_t$

        Update $\pi(\theta_{tc})$ using observed $(s_t, a_t, r_t, s_{t+1})$

        Update $\theta_{ta}^{(2)}$

    **end for**

---

*E. State-Action Space*

The state observation $s_t \in R^{45}$ of the quadruped robot primarily consists of the following seven components. Details

TABLE I QUADRUPED ROBOT OBSERVATIONS

| Observation | Dim. |
|---|---|
| Body angular velocity $\omega_b$ | 3 |
| Body frame gravity $\mathbf{g}$ | 3 |
| Joint positions $\mathbf{q}$ | 12 |
| Joint velocity $\dot{\mathbf{q}}$ | 12 |
| Command | 3 |
| Action | 12 |

are shown in the TABLE I.

The policy network is composed of a single neural network, which outputs joint position targets to the system. The state-action setup for the MIT Humanoid and XBot-L humanoid robots follows the configurations in [19] and [34]. The torque calculation is as follows:

$$\tau = K_p(\hat{\mathbf{q}} - \mathbf{q}) + K_d(\hat{\dot{\mathbf{q}}} - \dot{\mathbf{q}}) \quad (6)$$

For the Go2 robot, where $K_p = 40$, $K_d = 0.5$, and the setting

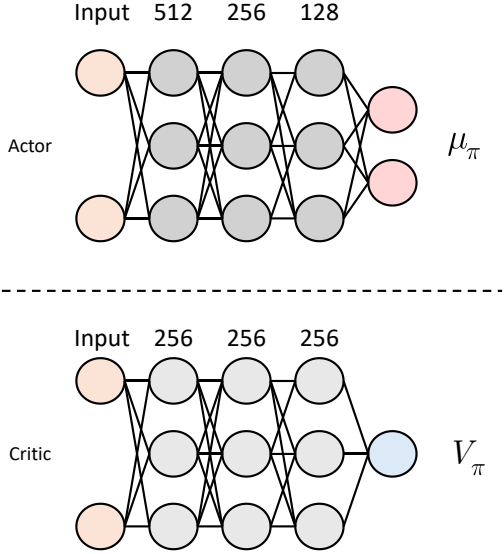

Fig. 3. The policy network structure of the legged robot.

is $\hat{\dot{\mathbf{q}}} = 0$.

*F. Policy Structure*

In the RL training for legged robots, we employed an actor-critic network, with both the actor and critic networks consisting of three layers, structured as MLPs. The detailed structure of the policy network can be seen in Fig. 3.

*F. Reward*

The MIT humanoid robot and XBot-L humanoid robot adopt the reward function settings from [19] and [34], respectively, while the Go2 reward function is designed to adjust the robot's actions through a system of rewards and penalties, aiming to enhance its stability, efficiency, and accuracy during task execution. These reward functions cover various aspects, including speed, posture, height, and torque. The specific reward functions for the Go2 robot are detailed in TABLE II. Here, $v_z$ is the linear velocity in the numerical direction, $\omega_x$ and $\omega_y$ are the angular velocities on the horizontal plane, $g_x$ and $g_y$ are the components of gravitational acceleration along the $x$ and $y$ axes, $\dot{q}_i^t$ is the joint acceleration of the $i$-th joint at time $t$, $a_i^t$ is the action of the $i$-th joint at time $t$, $\sigma$ is the smoothing factor, $\mathbf{c}_k$ is the commanded linear velocity, $v_k$ is the actual linear velocity, and $\mathbf{c}_1$ and $\mathbf{c}_2$ are the lower and upper limits of the joint position.

TABLE II Go2 TRAINING REWARDS

| Reward type | Weight | Function |
|---|---|---|
| Linear velocity tracking | 1.0 | $\exp\left(-\frac{1}{\sigma}\sum_{k\in(x,y)}(\mathbf{c}_k - v_k)^2\right)$ |
| Ang_velocity tracking | 0.5 | $\exp\left(-\frac{1}{\sigma}(\mathbf{c}_w - \omega_z)^2\right)$ |
| Z-axis velocity | -2 | $v_z^2$ |
| xy ang_velocity | -0.05 | $(\omega_x^2 + \omega_y^2)$ |
| Torques | $-2*e^{-4}$ | $\tau^2$ |
| Joint acceleration | $-2.5*e^{-7}$ | $\sum_i\left(\frac{\dot{q}_i^t - \dot{q}_i^{t-1}}{\Delta t}\right)^2$ |
| Joint position limits | -10 | $\min(-(\mathbf{q}-\mathbf{c}_1)) + \max(\mathbf{q}-\mathbf{c}_2)$ |
| Action change rate | -0.01 | $\sum_i(a_i^t - a_i^{t-1})^2$ |

TABLE III TRAINING ENVIRONMENT HYPERPARAMETERS

| Hyperparameter | Value |
|---|---|
| Value loss coefficient | 1.0 |
| Clipping $\epsilon$ | 0.2 |
| Learning rate | 1e-3(adaptive) |
| Discount factor | 0.998 |
| $\lambda$ | 0.95 |
| Activation | ELU |

## IV. EXPERIMENT

*A. Training and Implementation Details*

To evaluate the effectiveness of the proposed algorithm, we trained two humanoid robot models on the Isaac Gym [35] platform. The training was conducted on a computer equipped with an I9-14900k processor and an NVIDIA RTX 4090 GPU, using the PyTorch framework. In the simulation environment, control was executed through a PD controller, with a control frequency of 1kHz. The training involved 4096 agents, and the total training time was 3 hours. The design of the hyperparameters is detailed in TABLE III.

We trained the policy for the Go2 robot and transferred it to the XBot-L humanoid robot. The specific method involved freezing the first layer of the Go2's actor network and applying it to the XBot-L. The experimental results are shown in Fig. 4.

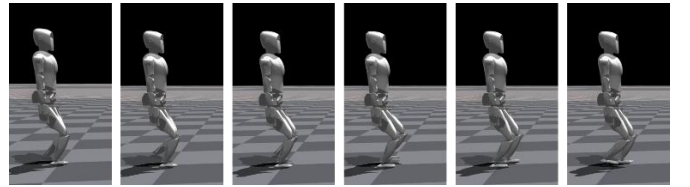

Fig. 4. The figure shows the motion sequence of the robot as it transitions from the Go2 to the XBot-L humanoid robot, displaying the robot's actions at each frame.

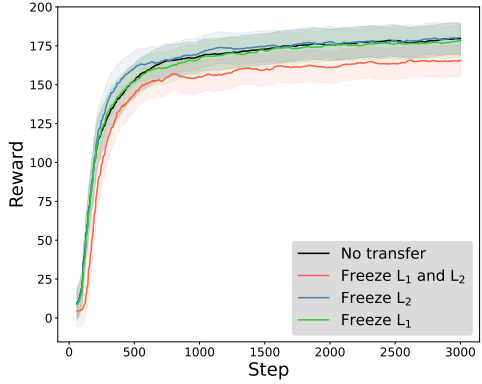

Fig. 5. The performance of the reward function with different fixed layers when migrating from Go2 to the XBot-L humanoid robot is shown. The black line represents no migration, the red line indicates both layers are frozen, the blue line represents the freezing of the later layer, and the green line represents the freezing of the earlier layer. The more parameters are frozen, the fewer parameters are available for fine-tuning.

### B. Comparison and Analysis

Due to the differences in observation and action dimensions among robots (different robots have varying numbers of joints), the input and output dimensions of the network differ accordingly. Therefore, the number of frozen layers can be chosen from [512, 256] or [256, 128]. In our experiment, we selected to freeze the [256, 128] layer, as it has fewer network parameters, allowing for more parameters to be fine-tuned. We define the [512, 256] layer as $L_1$ and the [256, 128] layer as $L_2$. Here, we compare the results of freezing $L_1$, freezing $L_2$, and freezing in both layers, along with the corresponding reward functions. The reward function images are shown in Fig. 5.

As shown in Fig. 5, compared to the black curve representing the method of training from scratch, freezing $L_1$ achieves comparable performance, and freezing $L_2$ results in a higher reward function than the non-migration approach. However, when both layers are frozen, the reward value is lower than that of the non-migration method. This indicates that prior knowledge can be effectively utilized during migration, but a certain amount of trainable parameters should be retained for fine-tuning. If too many parameters are frozen (such as freezing both layers), there will be insufficient remaining trainable parameters, making it difficult for the model to adapt to the new robot. In contrast, freezing only $L_2$ preserves more fine-tuning space, leading to a faster increase in the reward value and a higher final value. Additionally, to verify the time-saving effect of the proposed method, we recorded the time required for training. Training from scratch takes 3.01 hours, while the transfer learning model only takes 2.5 hours. In comparison, it saves about half an hour and can still reach the level of direct training.

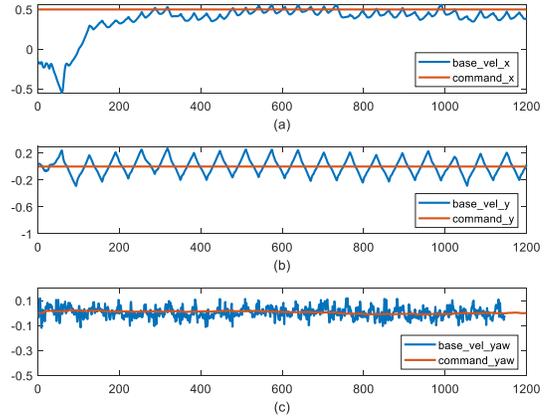

Fig. 6. (a) shows the commanded x-axis velocity and the actual velocity, (b) shows the commanded y-axis velocity and the actual velocity, and (c) shows the commanded yaw and the actual yaw.

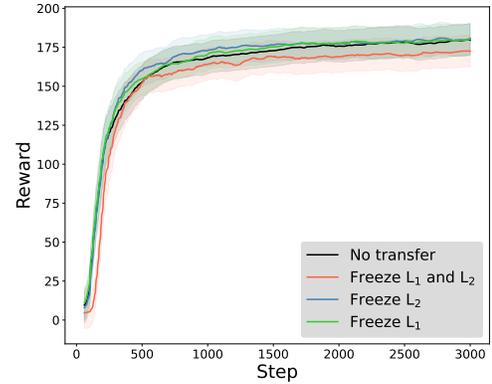

Fig. 7. The reward function performance from MIT humanoid robot to XBot-L humanoid robot with different fixed layers is shown, where the black curve represents no transfer, the red curve indicates both layers are frozen, the blue curve represents freezing the later layer, and the green curve represents freezing the earlier layer.

At the same time, we recorded the robot's base velocity in the x and y axes, as well as its yaw angle, as shown in Fig. 6. From the figure, it can be observed that for the x-axis velocity, although there was an initial deviation in the opposite direction, the velocity stabilized around the command value of 0.5 m/s thereafter. The fluctuations in the y-axis velocity indicate some side-to-side swaying during walking, but the amplitude is relatively small. The yaw angle performed well, further validating the reliability of our method.

We then conducted further transfer experiments between different humanoid robots. We trained a policy using the MIT humanoid robot and transferred it to the XBot-L humanoid robot, achieving the same experimental results. In comparison, due to the greater similarity between robots of the same type, the reward function performance was better than the results from transferring from the Go2 platform. The experimental results are shown in Fig. 7. This indicates that the PTRL method is more effective when applied to robots of the same type.

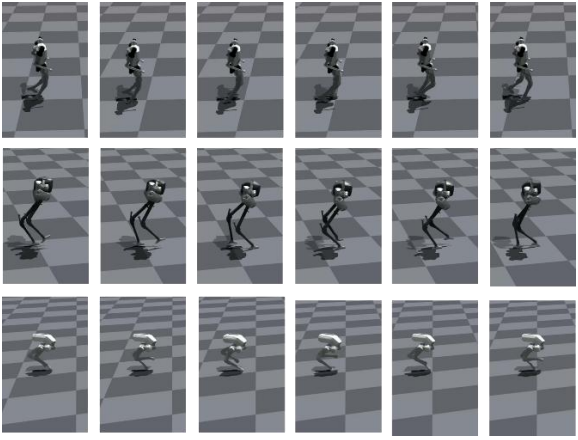

Fig. 8. The transfer from the XBot-L humanoid robot to legged robots is shown as follows: the top section shows the MIT humanoid, the middle section shows the Cassie humanoid robot and the bottom section shows the TRON1 robot.

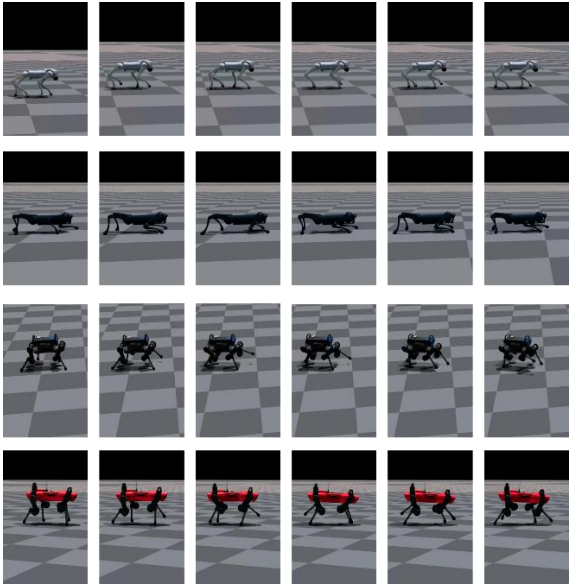

Fig. 9. The transfer from the XBot-L humanoid robot to the Go2, A1, Anymal_B, and Anymal_C robots is shown.

In the previous experiments, we validated the transfer of policies from the Go2 and MIT humanoid robots to the XBot-L humanoid robot. The next set of experiments extended to other legged robots, further testing the adaptability of the PTRL method across various robot platforms.

We transferred the policy trained on the XBot-L humanoid robot to other robot platforms, and the results are shown in Fig. 8 and Fig. 9. The results demonstrate that the PTRL method maintains good performance across multiple robot platforms, confirming its effectiveness in cross-platform transfer. Particularly, between quadruped and biped robots, the PTRL method effectively reduces training time while ensuring that the transferred policy performs similarly to the policy trained from scratch. These experiments further confirm the potential of the PTRL method for multi-platform and multi-robot system applications.

In the training of the eight robots mentioned above, we recorded a time comparison between training with transfer (WT) and without transfer (W/OT). Since domain randomization was applied to the XBot-L humanoid, Go2, and TRON1 robots, the number of training iterations and training time for these robots were longer compared to the others. The experimental results are shown in TABLE IV. From the results, we observed that for each robot, the training time was reduced by approximately 20 % on average, demonstrating the effectiveness of the proposed method. However, due to fewer training iterations for the Anymal robot, no significant difference was observed.

TABLE IV TRAINING TIME COMPARISON

|  | XBot-L | MIT | Cassie | Tron1 | Go2 | A1 | Anymal B | Anymal C |
|---|---|---|---|---|---|---|---|---|
| WT | 2.5h | 8min | 8min | 3h | 1.5h | 8min | 3min | 3min |
| W/OT | 3h | 12min | 11.5min | 3.38h | 1.75h | 12min | 3.3min | 3.3min |

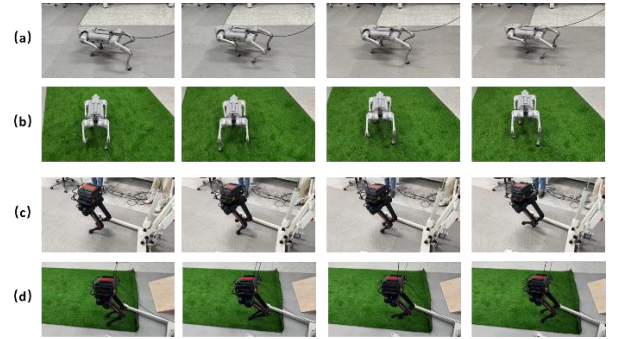

Fig. 10. (a) Go2 walking on flat ground; (a) Go2 walking on the grass; (c)TRON1 walking on flat ground; (d) TRON1 walking on the grass.

### C. Hardware Experiment

We successfully transferred the migrated and trained strategy to the real world using appropriate domain randomization. In experiments conducted on the Go2 and TRON1 platforms, the robots were tested walking on both flat ground and grass. The results demonstrated that both robots exhibited strong robustness and were able to walk stably on flat surfaces. The experimental images are shown in Fig. 10.

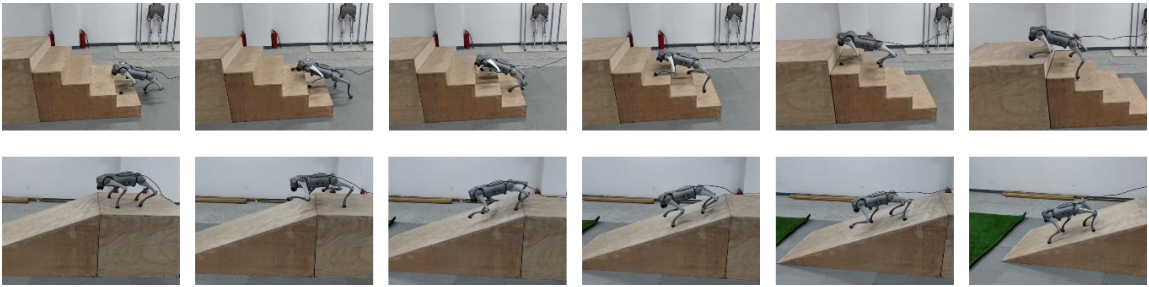

Fig. 11. Go2 walks on stairs and slopes.

Meanwhile, placing Go2 on a slope with a 20-degree incline and having it walk on stairs has resulted in good walking performance, as shown in Fig. 11.

V. CONCLUSION

This paper presents a novel Prior Transfer Reinforcement Learning (PTRL) framework, which effectively addresses the challenge of cross-robot policy transfer. By combining model transfer from deep learning with model fine-tuning techniques from natural language processing, the proposed approach not only reduces the number of training parameters and training time but also fully leverages existing prior knowledge. Simulation experiments conducted on quadruped robots and multiple different humanoid robot platforms validate the effectiveness of the proposed method. The results demonstrate that this approach can achieve walking performance comparable to direct reinforcement learning methods while reducing computational resource requirements and shortening training time. The PTRL method exhibits significant versatility.

Although the proposed method shows substantial advantages in cross-robot policy transfer, certain limitations remain in skill transfer and terrain adaptation, where the results did not fully meet expectations. Future research will focus on these areas, explicitly exploring strategies to improve skill transfer and enhance the transferability of policies in complex and diverse terrain conditions, with the aim of further improving the applicability and robustness of the method.